\documentclass{article}
\usepackage{ijcai26}

\usepackage{times}
\usepackage{soul}
\usepackage{url}
\usepackage[hidelinks]{hyperref}
\usepackage[utf8]{inputenc}
\usepackage[small]{caption}
\usepackage{graphicx}
\usepackage{amsmath}
\usepackage{amsthm}
\usepackage{booktabs}
\usepackage[switch]{lineno}
\usepackage{multirow}
\usepackage{tabularx}
\usepackage{subfigure}
\title{An Information-theoretic Propagation Denoising and Fusion Framework for \\ Fake News Detection}

\author{
Mengyang Chen$^{1,2}$
\and
Lingwei Wei$^{1,*}$
\and
Wei Zhou$^{1}$\And
Songlin Hu$^{1,2}$\\
\affiliations
$^1$Institute of Information Engineering, Chinese Academy of Sciences\\
$^2$School of Cyber Security, University of Chinese Academy of Sciences\\
\emails
\{chenmengyang, weilingwei, zhouwei, husonglin\}@iie.ac.cn
}

\begin{document}

\maketitle
\maketitle

\renewcommand{\thefootnote}{\fnsymbol{footnote}}
\footnotetext[1]{Corresponding author.}
\renewcommand{\thefootnote}{\arabic{footnote}}
\begin{abstract}

Incomplete propagation data significantly hinders robust fake news detection. Recent approaches leverage large language models to simulate missing user interactions via role-playing, thereby enriching propagation with synthetic signals. However, such propagation data is intrinsically unreliable, and directly fusing it can lead to biased representations and limited detection performance. In this paper, we alleviate the unreliability of synthetic propagation from the mutual information perspective and propose a novel information-theoretic propagation denoising and fusion (InfoPDF) framework to learn effective representations from both real and synthetic propagation.  
Specifically, we first generate attribute-specific synthetic propagation using large language models. Then we model each synthetic propagation graph as a probabilistic latent distribution to guide reliability-aware adaptive fusion with real propagation. During training, we design a mutual information-based objective to learn compressed and task-sufficient propagation representations.
It jointly suppresses noisy signals across attribute-specific synthetic propagation, maintains consistency between real and synthetic propagation representations, and ensures task sufficiency for fake news detection and attribute prediction. Experiments on three real-world datasets show that InfoPDF consistently achieves superior performance across various fake news detection tasks. Further analysis demonstrates that InfoPDF can estimate attribute-level reliabilities and learn more discriminative propagation representations.

\end{abstract}

\section{Introduction}

The proliferation of fake news poses a significant threat to societal stability. Existing detection approaches primarily rely on content features (e.g., articles and comments)~\cite{castillo2011information,ma2015detect,luvembe2023dual} or propagation patterns~\cite{bian2020rumor,wei-etal-2021-towards,jiang2025epidemiology}. Compared to content-based methods, propagation-based approaches generally yield superior performance by capturing rich structural features.
However, real-world propagation data is often incomplete due to platform limitations or malicious deletions~\cite{wei-etal-2021-towards,ma2022towards}, leading to information sparsity that compromises detection accuracy. While some works attempt to enhance robustness in modeling partial propagation~\cite{wei2022uncertainty,jiang2025epidemiology}, they remain constrained by the availability of initial real data. Recently, large language models (LLMs) have been leveraged to simulate missing interactions through attribute-mixed role-playing~\cite{nan2024let,wan2024dell}, offering a promising solution for incomplete propagation data.

\begin{figure}[t]
    \centering
    \includegraphics[width=0.95\linewidth]{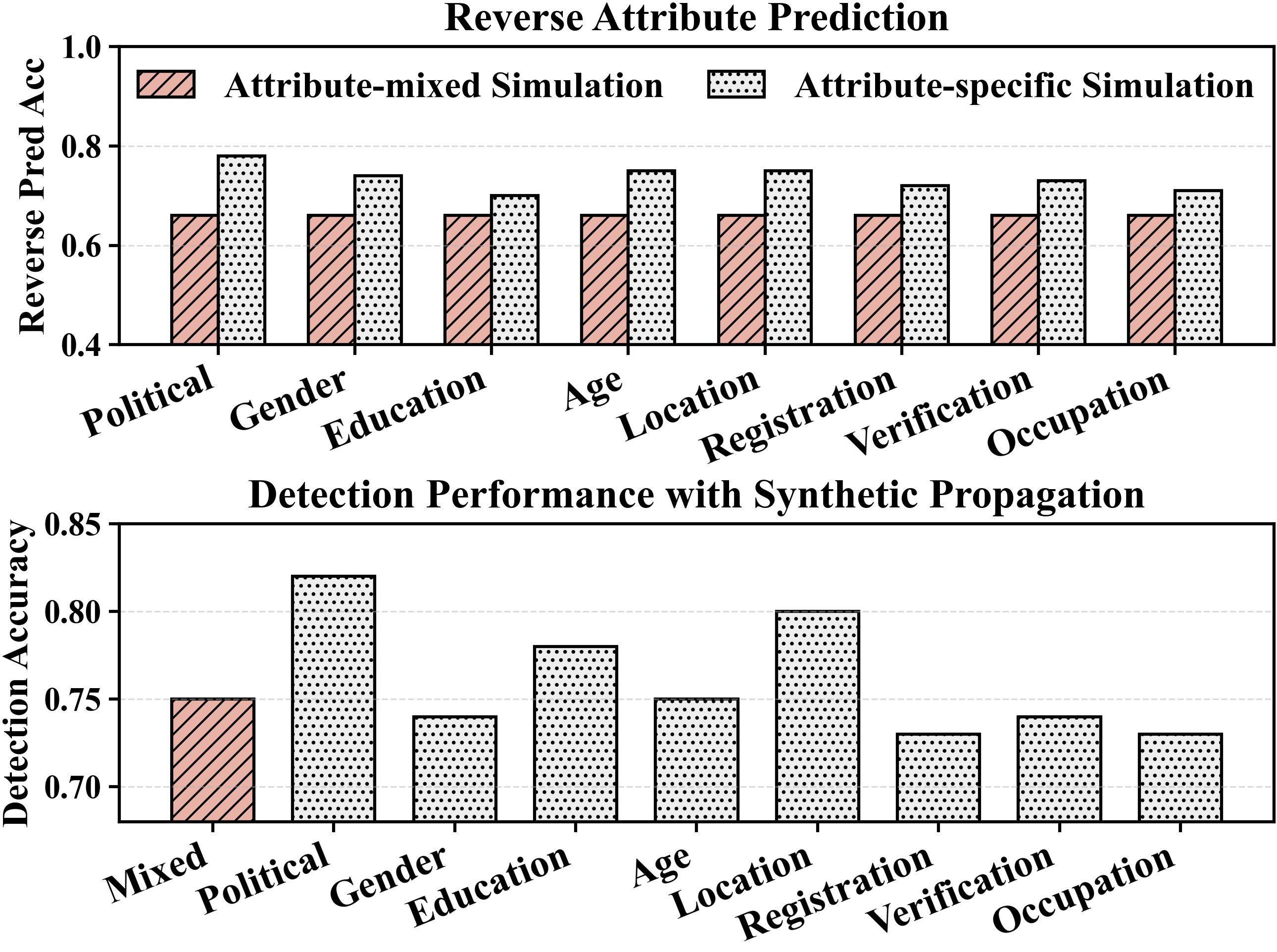}
    \caption{Preliminary analysis about different attributes and simulation settings on Twitter
    \protect\cite{lin2022detect}.
    Top: reverse attribute prediction accuracy based on the generated comments using GPT‑4 Turbo. 
    Bottom: fake news detection accuracy using synthetic propagation (GCN \protect\cite{kipf2016semi} as detector).
    }
    \label{fig:pre}
\end{figure}

However, these LLM-based simulation approaches overlook the inherent unreliability of synthetic data. Crucially, naive attribute-mixed prompting can further exacerbate this issue, i.e., LLMs may fail to fully capture the situational context of discussions, leading to generated interactions that contain hallucinations~\cite{ji2023towards}, as well as stylistic deviations and unnatural tones \cite{trott2023large}, which deviate from real-world discussions. Moreover, diverse user attributes (e.g., age and politics) impact fake news spread differently~\cite{shu2019role}. Mixing multiple attributes within a single prompt often yields inconsistent or contradictory signals. As a result, low-quality synthetic signals can contaminate informative ones, reducing the effective information available for learning and ultimately weakening detection.
As shown in a preliminary analysis in Figure~\ref{fig:pre}, performance varies significantly across different user attributes and simulation settings. This observation suggests that the reliability of LLM-generated synthetic propagation is attribute-dependent, motivating the explicit modeling of attribute-level reliabilities to distinguish informative signals from noise.

In this paper, we alleviate the unreliability of synthetic propagation from the mutual information perspective, and propose a novel information-theoretic propagation denoising and fusion (InfoPDF) framework to learn effective representations from both real and synthetic propagation.  
Specifically, InfoPDF integrates real and synthetic propagation for each attribute view to form a fused representation, which is modeled as a stochastic latent variable for uncertainty estimation. The estimated uncertainty is then converted into an attribute-level reliability score. It enables reliability-aware adaptive fusion to emphasize informative views while down-weighting noisy ones, thereby mitigating the negative impact of unreliable synthetic propagation on detection.
During training, we design a mutual information based objective to learn compressed and task-sufficient propagation representations. 
It consists of three complementary objectives, i.e., minimizing the mutual information between synthetic propagation and latent representations to suppress redundant noise across different attribute views;  maximizing the mutual information between real and synthetic propagation representations to reduce distributional discrepancy for propagation alignment; and maximizing the mutual information between latent representations and task labels to ensure task sufficiency for fake news detection and user attribute prediction.
Through this design, InfoPDF effectively denoises and fuses multi-view synthetic propagation, improving the robustness of detection under incomplete propagation scenarios.

We conduct experiments on three public real-world fake news datasets. 
{Experimental results demonstrate that InfoPDF consistently obtains superior performance in general detection as well as robust detection against noisy and incomplete scenarios.}
Further analysis reveals that our InfoPDF effectively estimates attribute-level reliability weights and extracts more discriminative propagation representations for fake news detection. 

Our contributions can be summarized as follows: 
1) We identify the unreliability of LLM-generated synthetic propagation for fake news detection, demonstrating that synthetic propagation can be noisy and its quality varies across attributes, which may reduce the effective information for detection.
2) We propose an LLM-generated propagation denoising and fusion framework to learn effective representations from both original and synthetic propagation. It denoises unreliable synthetic propagation via a compression-based denoising objective, and maintains the consistency between real and synthetic propagation representations as well as task sufficiency for fake news detection and user attribute prediction.
3) Experiments on three public datasets demonstrate the superiority of our proposed framework, validating its effectiveness and robustness in handling incomplete propagation scenarios.

\begin{figure*}[t]
    \centering
    \includegraphics[width=0.95\linewidth]{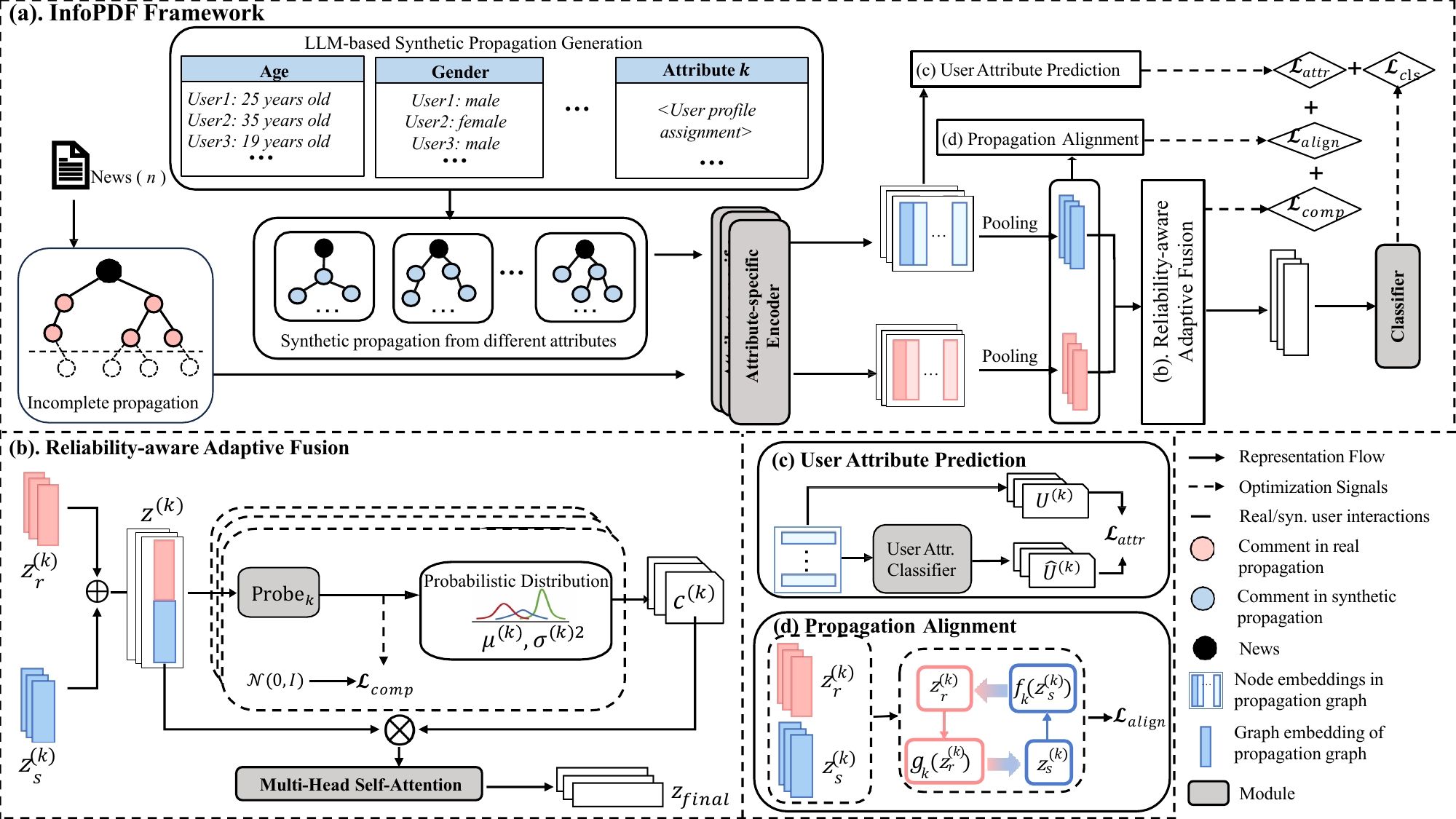}

\caption{Overview of the InfoPDF framework. InfoPDF generates attribute-specific synthetic propagation graphs, encodes them with incomplete real propagation, estimates reliability via probabilistic modeling, and performs reliability-aware fusion for fake news detection.}

    \label{fig:InfoPDF}
\end{figure*}

\section{Task Formulation}
{Fake news detection} aims to verify the authenticity of a news item, which can be formulated as a binary classification task.

Formally, each sample is represented by a graph $\mathcal{G}=({V},{E})$, where ${V}=\{n,c_1,\ldots,c_N\}$ denotes the news $n$ and its comments, and $\mathcal{E}$ denotes explicit interactions (e.g., reply).

Due to missing interactions in practice, the observed real propagation graph is often incomplete, denoted as $G_r=(V_r,E_r)$.
To enrich propagation signals, we construct $K$ attribute-specific synthetic graphs $\{G_s^{(k)}\}_{k=1}^{K}$, 
where each $G_s^{(k)}=(V_s^{(k)},E_s^{(k)})$ is conditioned on an attribute $a_k \in A=\{a_1,\ldots,a_K\}$. This allows precise assessment of each attribute's contribution.
The task goal can be formulated as,
    $$f: (G_r, \{G_s^{(k)}\}_{k=1}^K) \longrightarrow y, \quad y \in \{0, 1\}.
    \label{eq:objective}$$

\section{Methodology}
\label{sec:methodology}
In this section, we propose a principled InfoPDF framework to learn effective representations from both real and synthetic propagation for fake news detection.

\subsection{Overview of InfoPDF}
\label{sec:overview}

The architecture of InfoPDF is shown in Figure~\ref{fig:InfoPDF}.
We first leverage large language models (LLMs) to generate synthetic propagation based on each user attribute. 
We encode the incomplete real propagation graph and these attribute-specific synthetic propagation graphs by graph neural networks.
We model the fused representation as a stochastic latent variable to estimate the reliability of synthetic propagation for each attribute, which is converted to guide the adaptive fusion where informative views are emphasized and noisy ones are down-weighted.
During the training, we design a mutual-information based objective to jointly optimize the representations via minimizing the mutual information between synthetic propagation and latent representations to suppress redundant noise across different attribute views,  maximizing the mutual information between real and synthetic propagation representations to reduce distributional discrepancy for propagation alignment, and maximizing the mutual information between latent representations and task labels to ensure task sufficiency for fake news detection and user attribute prediction.
InfoPDF learns compressed and task-sufficient representations by exploiting both synthetic and real propagation for robust fake news detection.

\subsection{Reliability-Aware Graph Fusion Network}

Since the quality of synthetic propagation may vary across attributes, we model the reliability of multiple attribute-specific synthetic propagation graphs via variational estimation and use it to guide adaptive fusion.

\paragraph{LLM-based Synthetic Propagation Generation.}
Given a news item, we generate an attribute-specific synthetic propagation graph $G_s^{(k)}$ for each user attribute $a_k \in A$ using a frozen LLM, following the previous method~\cite{wan2024dell}. 
Beyond them, we consider eight user attributes including age, education~\cite{gaillard2021countering}, gender, occupation~\cite{lim2021local}, politics, registration time, verification status, and location~\cite{shu2019role}.

We then encode the incomplete real propagation graph $G_r$ and each synthetic graph $G_s^{(k)}$ into latent representations.
To ensure that real and synthetic views are comparable, we adopt a Siamese-style encoding strategy~\cite{chopra2005learning} to compute the fused representations.
Specifically, for the $k$-th attribute, we use a graph neural network encoder denoted as $\text{Enc}_k(\cdot)$ to compute the graph-level representations of real and synthetic propagation, i.e.,
\begin{align}
    \mathbf{z}_r^{(k)} &= \text{Pooling}(\text{Enc}_{k}(G_r)), \label{eq:encode_real} \\
    \mathbf{z}_s^{(k)} &= \text{Pooling}(\text{Enc}_{k}(G_s^{(k)})). \label{eq:encode_synth}
\end{align}
Then we build a unified representation for each attribute view by concatenating the real and synthetic branches, i.e.,
\begin{equation}
    \mathbf{z}^{(k)} = [\mathbf{z}_r^{(k)}; \mathbf{z}_s^{(k)}].
    \label{eq:unified_view}
\end{equation}
\paragraph{Reliability-aware Adaptive Fusion.}
To model the reliability of synthetic propagation, we introduce a distributional uncertainty estimator for each attribute-specific view $\mathbf{z}^{(k)}$. We implement it with a Gaussian distribution and sample from it via the reparameterization trick~\cite{kingma2013auto}:
\begin{align}
[\boldsymbol{\mu}^{(k)}, &\log \boldsymbol{\sigma}^{(k)2}]
= \text{Probe}_k(\mathbf{z}^{(k)}),  \\
\mathbf{m}^{(k)}
= &\boldsymbol{\mu}^{(k)}
+ \boldsymbol{\sigma}^{(k)} \odot \boldsymbol{\epsilon},
\quad \boldsymbol{\epsilon} \sim \mathcal{N}(0,\mathbf{I}),
\label{eq:reparam}
\end{align}
where $\text{Probe}_k(\cdot)$ is a lightweight MLP.

Since the standard deviation of the Gaussian posterior is given by $\boldsymbol{\sigma}^{(k)}$, we estimate the uncertainty of each attribute view analytically and convert it into a reliability weight:
\begin{align}
c^{(k)}
= 1&-\text{Softmax}(\frac{1}{d}\sum_{i=1}^{d}\boldsymbol{\sigma}_{i}^{(k)})_k,
\label{eq:credibility_score}
\end{align}
where $d$ is the feature dimension. Other probabilistic distributions, such as the Beta distribution, can also be used for reliability estimation~\cite{han2022trusted,ma-etal-2024-event}; we further discuss this in Section~\ref{beta}.

Finally, we perform reliability-weighted aggregation over attribute-specific views, followed by multi-head self-attention to obtain the final representation:
\begin{equation}
\mathbf{h}_{final} = \text{Attention}([c^{(1)}, \dots, c^{(K)}]\cdot[\mathbf{z}^{(1)},\dots,\mathbf{z}^{(K)}]^\text{T}). 
\label{eq:fusion}
\end{equation}
\paragraph{Fake News Detection.}
The final prediction of fake news is computed by:
\begin{equation}
    \mathbf{\hat{y}} = \text{Softmax}(\mathbf{W}_{cls} \mathbf{h}_{final} + \mathbf{b}_{cls}),
    \label{eq:pred}
\end{equation}
where $\mathbf{W}_{cls}$ and $\mathbf{b}_{cls}$ are learnable parameters.

\subsection{Training Objectives based on Mutual Information}
\label{sec:training_objectives}
We design a mutual information based objective to learn compressed and task-sufficient propagation representations. 

Our goal is to i) minimize the mutual information between the input of synthetic propagation and latent representations to suppress redundant noise across different attribute views, 
ii) maximize the mutual information between real and synthetic propagation representations to reduce distributional discrepancy for propagation alignment, and iii) maximize the mutual information between latent representations and task labels to ensure task sufficiency for both fake news detection and user attribute prediction.

Formally, let $X$ denote the input graph (i.e., the unified view of real and synthetic propagation), $Y$ is the ground-truth task label (i.e., graph-level label $Y_G$ for fake news detection or node-level label $Y_V$ for user attribution label), node-level and graph-level representations are $Z_V$ and $Z_G$. 
Given the Markov chain constraint $(Y_G, Y_V)\rightarrow X \rightarrow Z_V \rightarrow Z_G $, the overall objective of InfoPDF is formulated as:
\begin{equation}
    \max   \underbrace{I(Z_{G}; Y_G)+\alpha I(Z_V;Y_V)}_{\text{Task Sufficiency}} 
    - \gamma \underbrace{I( X; Z_G)}_{\text{Compression}}
    + \lambda \underbrace{I(Z^r_G; Z^s_G)}_{\text{Alignment}},
\end{equation}
where $Z^r_G$ and $Z^s_G$ are graph-level representations learned from real and synthetic propagation. $\alpha$, $\lambda$ and $\gamma$ are trade-off hyperparameters
controlling the weights of task supervision, propagation alignment, and information compression.

\paragraph{Task Sufficiency.}
We jointly optimize \textbf{fake news detection} and \textbf{user attribute prediction} via supervised learning objectives, which serve as variational lower bounds of the corresponding mutual information terms $I(Z_{G};Y_G)$ and $I(Z_V;Y_V)$.
First, maximizing $I(Z_G; Y_G)$ encourages the learned graph-level representation $Z_G$ to be maximally informative for fake news classification task, which is implemented by minimizing the binary cross-entropy loss,
\begin{equation}
    \mathcal{L}_{cls} = 
    - \log p_{\theta_{cls}}\!\left(Y_G | Z_G\right),
    \label{eq:cls_loss}
\end{equation}
where $\theta_{cls}$ refers to trainable parameters for fake news classifier.
Minimizing $\mathcal{L}_{cls}$ corresponds to maximizing the conditional log-likelihood $\log p(Y_G | Z_G)$ and thus a lower bound of $I(Z_G; Y_G)$.

Additionally, to encourage node-level representations to retain attribute-discriminative information, we introduce an auxiliary node-level attribute prediction objective to maximize $I(Z_V; Y_V)$.
We minimize the following negative log-likelihood loss, i.e.,
\begin{equation}
\small
    \mathcal{L}_{attr} = 
    \frac{1}{K}\sum_{k=1}^{K}\frac{1}{\log|\mathcal{C}_k|}
\left(-\log p_{\theta^{(k)}_{attr}}\!\left(Y_V^{(k)} \mid Z_V^{(k)}\right)\right),
    \label{eq:attr_loss}
\end{equation}
where $|\mathcal{C}_k|$ is the number of classes for attribute $a_k$.
$\theta_{attr}^{(k)}$ refers to trainable parameters for the $k$-th attribution classifier.
We use the log-scale factor to balance attribute spaces with different labels.
Minimizing $\mathcal{L}_{attr}$ maximizes the conditional likelihood of attribute labels given node representations, serving as a variational lower bound for $I(Z_V; Y_V)$.

\paragraph{Information Compression.} 
To suppress redundant information introduced by LLM-generated synthetic propagation, we adopt the variational information bottleneck~\cite{alemi2017deep} to minimize the mutual information between the latent representation and the input view.
Specifically, we approximate each attribute-specific view with a variational Gaussian posterior
$q(\mathbf{m}^{(k)} | \mathbf{z}^{(k)})=\mathcal{N}(\boldsymbol{\mu}^{(k)}, \boldsymbol{\sigma}^{2(k)})$,
and impose a standard normal prior $p(\mathbf{m})=\mathcal{N}(\mathbf{0}, \mathbf{I})$.
The mutual information $I(X;Z)$ can be upper bounded by the Kullback--Leibler (KL) divergence between the posterior and the prior.
That is, we minimize the following compression loss, 
\begin{equation}
    \mathcal{L}_{comp} = \sum_{k=1}^{K} 
    D_{KL}\big(q(\mathbf{m}^{(k)}|\mathbf{z}^{(k)}) \,\|\, p(\mathbf{m})\big).
    \label{eq:comp_loss}
\end{equation}
Minimizing $\mathcal{L}_{comp}$ reduces the upper bound of $I(X;Z)$, encouraging accurate reliability estimation of each attribute-specific view by suppressing task-irrelevant noise.

\paragraph{Propagation Alignment.}
Since directly estimating mutual information is intractable, we adopt a deterministic alignment objective as a surrogate. Specifically, we introduce a bi-directional alignment loss function to reduce the gap between real and synthetic propagation representations.
The intuition is that when two views share high mutual information, one view should be predictive of the other~\cite{yu2025incomplete}.
For each attribute view $k$, we employ two lightweight predictors $f_k(\cdot)$ and $g_k(\cdot)$ (implemented as MLPs) to reconstruct one representation from the other. The loss function is defined as,
\begin{equation}
\small
    \mathcal{L}_{align} = \frac{1}{K}\sum_{k=1}^{K} \left( \|f_k(\mathbf{z}_s^{(k)}) - \mathbf{z}_r^{(k)}\|^2_2 + \|g_k(\mathbf{z}_r^{(k)}) - \mathbf{z}_s^{(k)}\|^2_2 \right).
    \label{eq:loss_align}
\end{equation}

Overall, the training loss of InfoPDF is defined as:
\begin{equation}
    \mathcal{L} = \mathcal{L}_{cls} + \alpha \mathcal{L}_{attr} + \lambda \mathcal{L}_{align} + \gamma \mathcal{L}_{comp},
    \label{eq:final_loss}
\end{equation}
where $\alpha$, $\lambda$, and $\gamma$ are trade-off hyperparameters.

\subsection{LLM-free Inference}
\label{llm_free}

Inspired by efficient LLM-free deployment~\cite{hu2024bad}, we provide an inference mode without online propagation simulation.
The trained predictor $g_k$ estimates the synthetic attribute representation as
\begin{equation}
\label{eq:completion_pred}
\hat{z}_s^{(k)} = g_k(z_r^{(k)}).
\end{equation}
Then, $\hat{{z}}^{(k)}=[z_r^{(k)};\hat{z}_s^{(k)}]$ is fed into reliability-aware fusion to obtain $h_{final}$ for detection.
\newcommand{\std}[1]{{\scriptsize$\pm$#1}}

\begin{table*}[t]
    \centering
    \small
    \begin{tabular}{l|l|cc|cc|cc}
        \toprule 
        \multicolumn{2}{c|}{\multirow{2}{*}{\bf Method}} 
        & \multicolumn{2}{c|}{\bf Twitter} 
        & \multicolumn{2}{c|}{\bf CED} 
        & \multicolumn{2}{c}{\bf PHEME} \\ 
        \multicolumn{2}{c|}{}  
        & Acc & F1 
        & Acc & F1 
        & Acc & F1 \\ 
        \cmidrule{1-8}
        
        \multirow{13}{*}{\rotatebox{90}{Content-based}} 
          & {BERT} & 71.12\std{2.43} & 70.86\std{2.71} & 88.44\std{0.96} & 88.43\std{0.89} & 82.96\std{0.61} & 80.57\std{0.82} \\ 
          & {DeBerta} & 75.43\std{1.92} & 74.47\std{1.87} & 89.97\std{0.73} & 89.32\std{0.79} & 82.54\std{1.09} & 80.24\std{1.21} \\ 
          & {EANN} & 72.84\std{1.68} & 72.41\std{1.79} & 89.73\std{0.76} & 89.69\std{0.72} & 83.27\std{0.41} & 81.07\std{0.27} \\ 
          & {dEFEND} & 75.12\std{1.41} & 73.56\std{1.47} & 90.12\std{0.63} & 89.98\std{0.71} & 83.24\std{0.84} & 82.45\std{0.91} \\ 
          & {DualEmo} & 73.27\std{1.57} & 73.01\std{1.66} & 91.21\std{0.71} & 91.20\std{0.71} & 83.60\std{0.88} & 83.34\std{0.86} \\ 
          \cmidrule{2-8}

          & {LLM$_\text{text}$} & 56.00\std{4.23} & 54.60\std{4.41} & 48.08\std{2.53} & 37.22\std{2.74} & 33.74\std{3.35} & 28.64\std{3.52} \\ 
          & {ARG} & 78.66\std{2.35} & 78.59\std{2.42} & 92.92\std{0.31} & 92.88\std{0.35} & 79.74\std{0.51} & 77.52\std{0.42} \\ 
          & {LLM$_\text{comment}$} & 60.75\std{3.64} & 60.47\std{3.78} & 80.10\std{1.41} & 80.05\std{1.38} & 53.51\std{2.41} & 52.77\std{2.58} \\ 
          & {TELLER} & 64.68\std{3.50} & 62.27\std{6.13} & 74.19\std{0.71} & 73.88\std{0.53} & 60.88\std{0.07} & 38.65\std{0.38} \\
          & {GenFEND} & & & & & & \\ 
          & {\quad w/ BERT} & 78.26\std{1.38} & 75.64\std{1.52} & 90.23\std{0.59} & 90.04\std{0.63} & 83.84\scriptsize{±0.81} &81.23\scriptsize{±0.87}  \\ 
          & {\quad w/ dEFEND} & 82.25\std{0.74} & 82.04\std{0.81} & 93.23\std{0.24} & 92.96\std{0.27} & 85.07\scriptsize{±0.39} &84.66\scriptsize{±0.45} \\
          & {CAMERED} & 82.85\std{0.65} & 82.52\std{0.71} & 93.58\std{0.22} & 93.25\std{0.25} & 85.44\scriptsize{±0.36} & 85.02\scriptsize{±0.42} \\
        \midrule

        \multirow{23}{*}{\rotatebox{90}{Propagation-based}}    
          & {GCN} & 78.02\std{2.91} & 77.67\std{3.08} & 91.59\std{1.12} & 91.17\std{1.08} & 80.59\std{0.30} & 75.55\std{1.05} \\ 
          & {GAT} & 79.31\std{2.73} & 78.86\std{2.87} & 88.23\std{1.29} & 87.68\std{1.31} & 80.41\std{0.46} & 77.13\std{1.25} \\ 
          & {GraphSAGE} & 81.03\std{2.58} & 79.75\std{2.69} & 89.26\std{1.19} & 88.92\std{1.26} & 79.03\std{1.29} & 75.70\std{0.52} \\ 
          & {BiGCN} & 82.76\std{1.56} & 82.71\std{1.49} & 91.74\std{0.69} & 91.63\std{0.74} & 82.70\std{0.35} & 80.28\std{0.88} \\ 
          & {UPFD} & 79.74\std{1.89} & 79.02\std{2.01} & 88.79\std{1.02} & 88.38\std{0.98} & 81.05\std{0.41} & 77.53\std{1.12} \\ 
          & {EBGCN} & 83.19\std{0.91} & 82.60\std{0.98} & 93.96\std{0.36} & 93.76\std{0.39} & 83.89\std{0.28} & 80.72\std{0.52} \\ 
          & {UPSR} & 83.62\std{0.87} & 83.22\std{0.94} & 93.36\std{0.33} & 92.96\std{0.36} & 82.95\std{0.31} & 82.95\std{0.37} \\ 
          & {RAGCL} & 84.42\std{0.72} & 84.15\std{0.78} & 94.10\std{0.29} & 93.95\std{0.32} & 84.24\std{0.25} & 83.58\std{0.41} \\
          \cmidrule{2-8}

          & {LLM$_\text{propagation}$} & 48.84\std{4.42} & 46.76\std{4.59} & 71.15\std{1.91} & 67.32\std{2.04} & 52.14\std{2.64} & 52.04\std{2.78} \\ 

          & {DELL} & & & & & & \\ 
          & {\quad - $single$} & 80.17\std{1.62} & 79.75\std{1.54} & 92.12\std{0.71} & 91.69\std{0.68} & 82.60\std{0.38} & 80.75\std{0.73} \\ 
          & {\quad - $vanilla$} & 80.43\std{0.58} & 80.32\std{0.59} & 92.23\std{0.62} & 91.75\std{0.66} & 82.52\std{0.42} & 80.60\std{0.68} \\ 
          & {\quad - $confidence$} & 78.97\std{1.69} & 78.10\std{1.73} & 91.03\std{0.68} & 90.10\std{0.72} & 83.03\std{0.45} & 81.08\std{0.76} \\ 
          & {\quad - $selective$} & 81.22\std{1.43} & 81.02\std{1.51} & 92.28\std{0.59} & 91.32\std{0.65} & 83.29\std{0.39} & 82.75\std{0.62} \\ 

          & {EIN} & 84.33\std{0.64} & 83.97\std{0.71} & 94.25\std{0.26} & 94.08\std{0.29} & 85.27\std{0.23} & 84.46\std{0.38} \\
          
          & {\bf InfoPDF (Ours)} & & & & & & \\ 
            & {\quad - GCN} 
            & 85.63$^*$\std{0.39} & 85.32$^*$\std{0.40} 
            & 96.36$^*$\std{0.07} & 96.24$^*$\std{0.06} 
            & 87.65$^*$\std{0.24} & 85.47\std{0.26} \\ 
            
            & {\quad - GAT} 
            & 85.43$^*$\std{0.32} & 85.15$^*$\std{0.34} 
            & \textbf{97.40}$^*$\std{0.18} & \textbf{97.28}$^*$\std{0.18} 
            & \textbf{88.17}$^*$\std{0.21} & 85.96$^*$\std{0.28} \\ 
            
            & {\quad - GraphSAGE} 
            & \textbf{86.90}$^*$\std{0.34} & \textbf{86.51}$^*$\std{0.35} 
            & 97.25$^*$\std{0.18} & 97.13$^*$\std{0.19} 
            & 88.11$^*$\std{0.22} & \textbf{86.11}$^*$\std{0.25} \\ 
          \cmidrule{2-8}

          & {\bf InfoPDF-Inference (Ours)} & & & & & & \\ 
          & {\quad - GCN} & 84.78\std{0.43} & 84.45\std{0.46} & 94.48\std{0.12} & 94.31\std{0.13} & 85.42\std{0.28} & 84.68\std{0.31} \\
          & {\quad - GAT} & 84.62\std{0.36} & 84.28\std{0.39} & 94.55\std{0.21} & 94.38\std{0.22} & 85.58\std{0.24} & 84.85\std{0.27} \\ 
          & {\quad - GraphSAGE} & 84.95\std{0.38} & 84.72\std{0.41} & 94.71\std{0.20} & 94.56\std{0.21} & 85.71\std{0.25} & 84.92\std{0.28} \\
        \bottomrule 
    \end{tabular}
    \caption{Results (\%) for fake news detection on three datasets. The best results are in \textbf{boldface}. InfoPDF-Inference represents the LLM-free inference mode of InfoPDF. $^*$ represents statistical significance over state-of-the-art baselines under the $t$-test ($p<0.05$).}
    \label{tab:main_results}
\end{table*}

\section{Experiments}

\subsection{Experimental Setups}

\paragraph{Datasets.} We conduct experiments on three public datasets, i.e., {Twitter}~\cite{lin2022detect}, {PHEME}~\cite{zubiaga2016learning}, and {CED}~\cite{song2019ced}.

\paragraph{Comparison Methods.}
We compare InfoPDF with 11 content-based methods including BERT~\cite{devlin2018bert} and DeBerta~\cite{He2020DeBERTaDB}, EANN~\cite{wang2018eann}, dEFEND~\cite{shu2019defend}, DualEmo~\cite{zhang2021mining}, ARG~\cite{hu2024bad}, GenFEND~\cite{nan2024let}, TELLER~\cite{liu-etal-2024-teller}, CAMERED~\cite{wang2025collaboration}, LLM$_\text{text}$, and LLM$_\text{comment}$~\cite{chen2025explore}, and 11 propagation-based methods including  GCN~\cite{kipf2016semi}, GAT~\cite{velivckovic2017graph}, GraphSAGE~\cite{hamilton2017inductive}, Bi-GCN~\cite{bian2020rumor}, EBGCN~\cite{wei-etal-2021-towards}, UPSR~\cite{wei2022uncertainty}, RAGCL~\cite{cui2024propagation}, UPFD~\cite{dou2021user}, DELL~\cite{wan2024dell}, EIN~\cite{jiang2025epidemiology}, and LLM$_\text{propagation}$ \cite{chen2025explore}. 

\paragraph{Evaluation Metrics.}
We evaluate with 
the accuracy score (Acc) and  macro-F1 score (F1).

\paragraph{Implementation Details.}
All experiments are implemented on a single NVIDIA Tesla V100 GPU. The size of synthetic propagation graphs is fixed to 30 nodes (1 root news node and 29 comment nodes). For the generation of synthetic propagation, we employ \textit{Mistral-7B-Instruct-v0.2} for Twitter and PHEME, and \textit{Qwen1.5-7B-Chat} for CED.

\begin{figure}[t]
\centering
{\includegraphics[width=\linewidth]{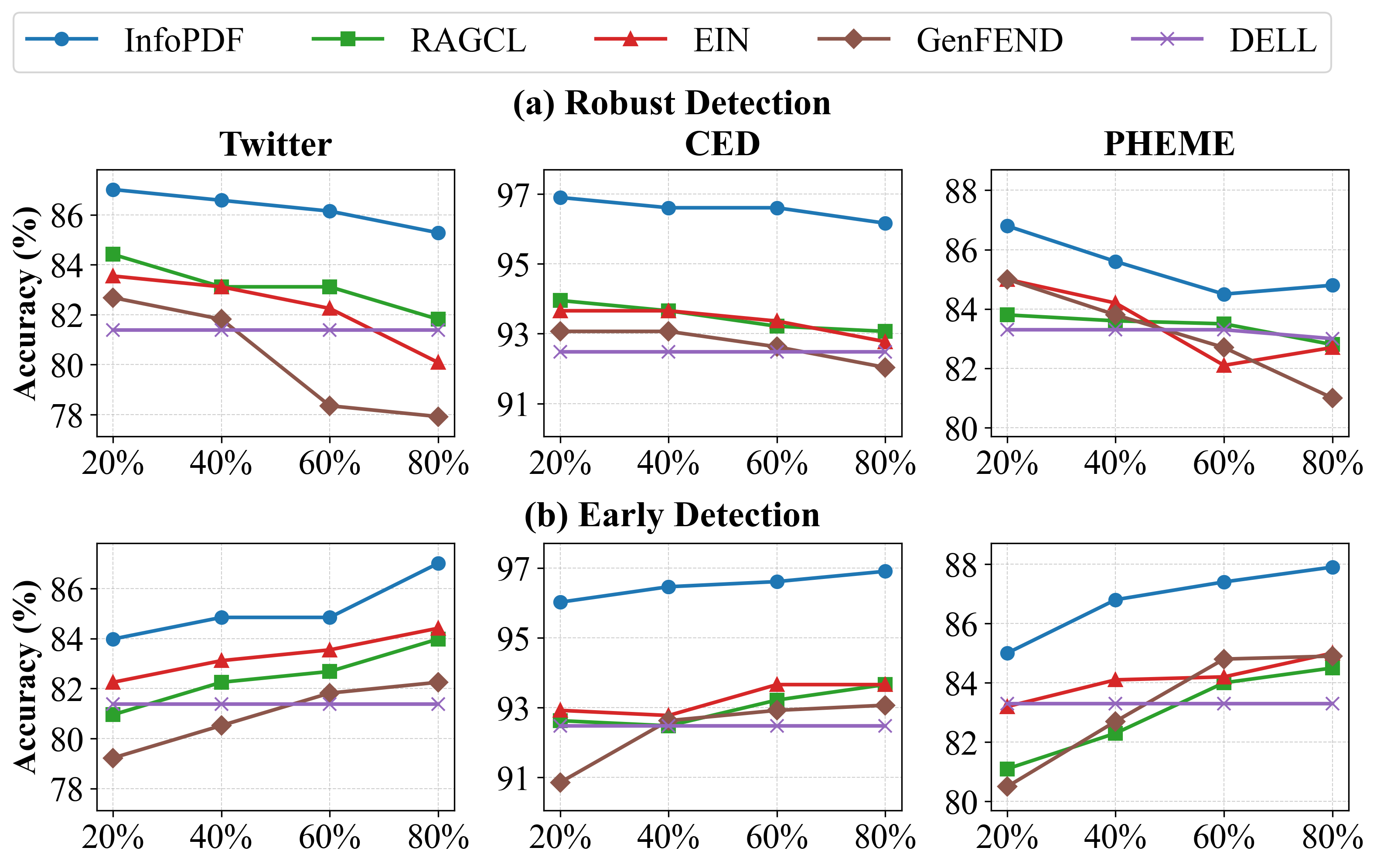}}
\caption{Results (\%) of robust and early fake news detection on two incomplete propagation scenarios.}
\label{fig:robust}
\end{figure}

\subsection{Main Results}
\label{main_results}
We evaluate the effectiveness of InfoPDF across different detection scenarios.

\paragraph{General Detection.}
Table~\ref{tab:main_results} reports the performance comparison on three public datasets.
Overall, InfoPDF achieves strong and generally superior performance across datasets and GNN backbones, demonstrating its effectiveness and strong generalization under incomplete propagation.
For instance, InfoPDF improves the GraphSAGE backbone by +5.87\% accuracy scores on Twitter. 
Moreover, LLM-free variant InfoPDF-Inference remains competitive with strong baselines while reducing the deployment cost by avoiding online LLM-based propagation simulation.
This validates that InfoPDF can distill the benefit of LLM-generated synthetic propagation into an efficient LLM-free detector, making it more practical for real-world deployment.

\paragraph{Robust Detection.}
Figure~\ref{fig:robust}(a) shows the results of InfoPDF and baselines for robust detection.
Specifically, we simulate noisy propagation settings by randomly removing edges and masking node features with ratios of 20\%, 40\%, 60\%, and 80\%.
From the results, InfoPDF shows smaller performance degradation, exhibiting stronger robustness towards different incomplete propagation settings.

\paragraph{Early Detection.}
Figure~\ref{fig:robust}(b) shows the results for early detection on three datasets.
We preserve the temporal order of propagation and use only partial propagation observed at early time steps for prediction.
InfoPDF achieves competitive performance even when only limited early propagation is available, showing its effectiveness in early detection.

\begin{table}[t]
\centering
\small
\resizebox{\linewidth}{!}{
\begin{tabular}{l|cc|cc|cc}
\toprule
\multicolumn{1}{c|}{\multirow{2}{*}{{\bf Method}}} & \multicolumn{2}{c|}{\bf Twitter} & \multicolumn{2}{c|}{\bf CED} & \multicolumn{2}{c}{\bf PHEME} \\
 & Acc & F1 & Acc & F1 & Acc & F1 \\
\midrule
\textbf{InfoPDF} & \textbf{86.90} & \textbf{86.51} & \textbf{97.25} & \textbf{97.13} & \textbf{88.11} & \textbf{86.11} \\
\midrule
\ \ w/o RP        & 84.34 & 84.10 & 94.05 & 93.89 & 86.21 & 85.97 \\
\ \ w/o Attr       & 84.70 & 84.23 & 94.79 & 94.54 & 86.04 & 85.64 \\
\ \ w/o RW            & 83.84 & 83.33 & 94.39 & 94.25 & 85.78 & 85.14 \\
\ \ w/o Attention      & 84.48 & 84.14 & 94.54 & 94.36 & 85.86 & 85.24 \\
\midrule
\ \ w/o $\mathcal{L}_{align}$ & 83.20 & {82.93} & 93.95 & {93.85} & 85.34 & 84.91 \\
\ \ w/o $\mathcal{L}_{comp}$    & 84.42 & 84.17 & 94.40 & 94.16 & 85.87 & 85.48 \\
\ \ w/o $\mathcal{L}_{attr}$  & 84.99 & 84.83 & 94.69 & 94.32 & 86.12 & 85.78 \\
\bottomrule
\end{tabular}
}
\caption{Ablation results on three datasets. 
The best performance is highlighted in bold. We apply GraphSAGE as the GNN backbone.
}
\label{tab:ablation}
\end{table}
\subsection{Ablation Study}\label{ablation}

We conduct ablation studies to examine the contribution of each component in InfoPDF. The ablation results are shown in Table~\ref{tab:ablation}.
For propagation modeling for fake news detection, removing the incomplete real propagation (\emph{w/o} RP) consistently degrades performance, showing the necessity of real propagation data.
Replacing attribute-specific simulation with attribute-mixed generation (\emph{w/o} Attr) leads to clear drops, indicating that modeling attributes separately yields more reliable synthetic signals.
Moreover, removing either reliability-aware weighting (\emph{w/o} RW) or multi-head self-attention fusion causes performance degradation, verifying their effectiveness for detection.
For ablation variants of training objectives, removing the propagation alignment objective (\emph{w/o $\mathcal{L}_{align}$}) results in  a large performance drop,
highlighting the importance of reducing the representation gap between real and synthetic propagation.
Removing the compression regularization (\emph{w/o $\mathcal{L}_{comp}$}) also consistently harms performance,
showing that the compression helps filter redundant features in representations.
Finally, removing attribute supervision (\emph{w/o $\mathcal{L}_{attr}$}) leads to further degradation,
demonstrating that preserving the sufficiency of representations for user attributes is beneficial for detection.

\subsection{Further Analysis}

We provide more analysis on the reliability of synthetic propagation,  representation quality under different fusion strategies, reliability estimation, and hyperparameter sensitivity.

\begin{figure}[t]
    \centering
    \includegraphics[width=\linewidth]{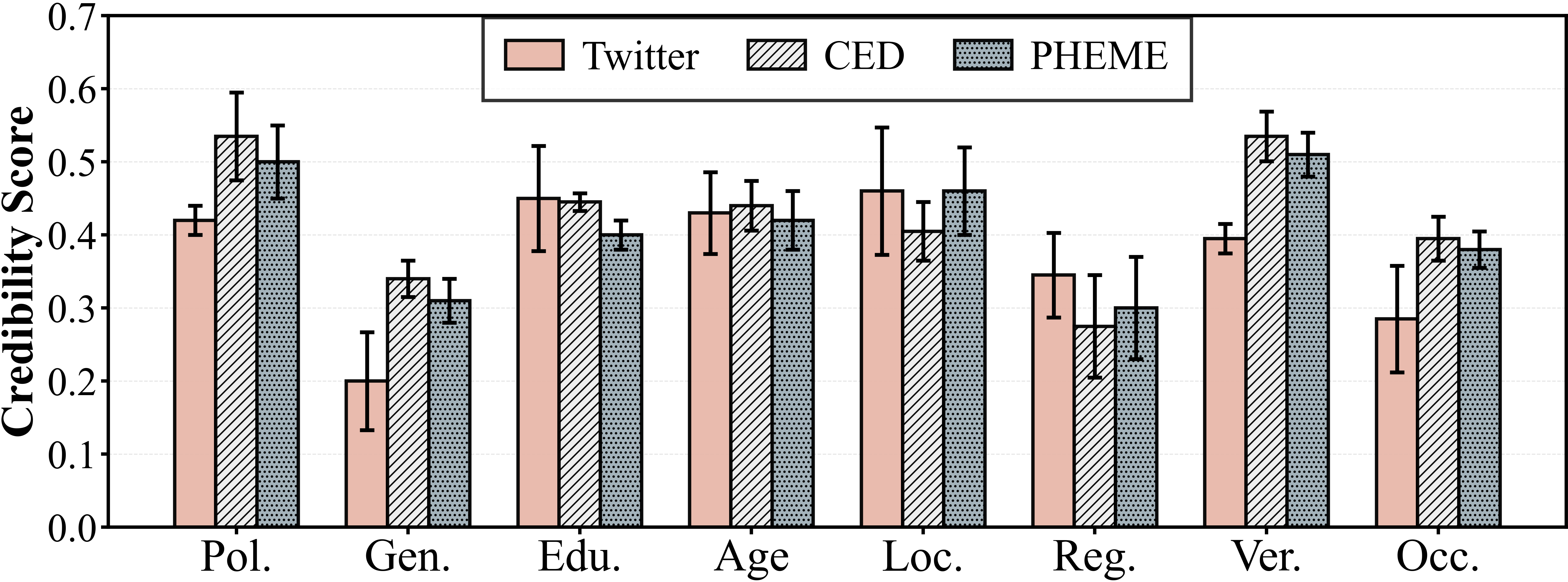}
    \caption{Attribute-level reliability distributions with user variability across three datasets.}
    \label{fig:credibility_distribution}
\end{figure}

\begin{table}[t]
\centering
\small
\resizebox{\linewidth}{!}{
\begin{tabular}{c|l|ccc|ccc}
\toprule
\multirow{2}{*}{\textbf{Dataset}} & \multicolumn{1}{c|}{\multirow{2}{*}{\textbf{Method}}} & \multicolumn{3}{c|}{\textbf{Supervised Metrics}} & \multicolumn{3}{c}{\textbf{Unsupervised Metrics}} \\ 
 & & ARI$\uparrow$ & NMI$\uparrow$ & FMI$\uparrow$ & SC$\uparrow$ & CHI$\uparrow$ & DBI$\downarrow$ \\ 
\midrule
\multirow{3}{*}{Twitter} 
 & RP & 30.14 & 23.51 & 65.08 & 0.24 & 85.30 & 1.52 \\ 
  & DF & 39.34 & 30.87 & 69.56 & 0.23 & 77.60 & 1.59 \\ 
 & \textbf{InfoPDF} & \textbf{43.82} & \textbf{34.67} & \textbf{71.79} & \textbf{0.56} & \textbf{443.54} & \textbf{0.59} \\ 
\midrule
\multirow{3}{*}{CED} 
 & RP & 65.28 & 57.60 & 82.70 & 0.56 & 1887.89 & 0.50 \\ 
  & DF & 60.13 & 50.60 & 80.10 & 0.38 & 543.31 & 1.06 \\ 
 & \textbf{InfoPDF} & \textbf{89.09} & \textbf{81.56} & \textbf{94.54} & \textbf{0.67} & \textbf{2564.49} & \textbf{0.43} \\ 
\midrule
\multirow{3}{*}{PHEME} 
 & RP & 41.47 & 28.77 & 74.58 & 0.11 & 139.66 & 2.53 \\ 
  & DF & 40.36 & 28.71 & 73.07 & 0.10 & 133.82 & 2.77 \\ 
 & \textbf{InfoPDF} & \textbf{51.12} & \textbf{39.52} & \textbf{77.64} & \textbf{0.49} & \textbf{1699.84} & \textbf{0.74} \\ 
\bottomrule
\end{tabular}
}
 \caption{Representation evaluation results (\%) against the propagation data from different strategies. Adjusted Rand Index (ARI), Normalized Mutual Information (NMI), and Fowlkes-Mallows Index (FMI) evaluate the accuracy of clustering. 
                 Silhouette Coefficient (SC), Calinski-Harabasz Index (CHI), and Davies-Bouldin Index (DBI) evaluate the separation and compactness of clustering. SC, CHI, and DBI are evaluated based on K-Means, and we define the number of clusters K as the true number of categories.
                }    \label{tab:representation}   
\end{table}

\paragraph{Reliability Analysis of Synthetic Propagation.}
\label{credibility_distribution}
Figure~\ref{fig:credibility_distribution} visualizes the attribute-view reliability scores estimated by the Beta variant of InfoPDF across three datasets.
Overall, we observe clear differences in the estimated reliability across attribute views, indicating that LLM-generated propagation signals exhibit varying levels of consistency and usefulness.
Moreover, the reliability scores also vary across instances within the same attribute view, suggesting that InfoPDF performs instance-wise reliability estimation rather than relying on static attribute-level priors.
These results support the effectiveness of reliability-aware fusion in adaptively leveraging informative synthetic views.

\paragraph{Representation Quality Analysis.}
\label{Representation Evaluation}
We analyze the quality of representations learned by InfoPDF and several variants including:
1) Real Propagation (RP), which encodes only the incomplete real propagation; and 2) Direct Fusion (DF), which fuses the real propagation representation with synthetic representations directly.
Following existing studies \cite{hu-etal-2023-supervised}, we compute supervised clustering metrics (ARI, NMI, FMI) and unsupervised clustering metrics (SC, CHI, DBI) to evaluate the representation quality. 
Table~\ref{tab:representation} reports results of our method and comparison variants.
The more accurate supervised clustering results demonstrate the class discrimination capability of InfoPDF.
Better results in terms of unsupervised metrics indicate the high structural quality of representations learned by our method, reflected in compact and well-separated clustering patterns. 

\paragraph{Reliability Estimation with Different Distributions.}
\label{beta}

\begin{table}[t] 
\centering

\resizebox{\linewidth}{!}{ 
\begin{tabular}{l|cc|cc|cc} 
\toprule 
\multicolumn{1}{c|}{\multirow{2}{*}{\textbf{Method}}} & 
\multicolumn{2}{c|}{\textbf{Twitter}} & \multicolumn{2}{c|}{\textbf{CED}} & \multicolumn{2}{c}{\textbf{PHEME}} \\
& Acc & F1 & Acc & F1 & Acc & F1 \\ \midrule \textbf{InfoPDF (Gaussian)} & & & & & & \\ 
\quad w/ GCN & 85.63 & 85.32 & 96.36 & 96.24 & 87.65 & 85.47 \\ 
\quad w/ GAT & 85.43 & 85.15 & \textbf{97.40} & \textbf{97.28} & \textbf{88.17} & 85.96 \\ 
\quad w/ SAGE & \textbf{86.90} & \textbf{86.51} & 97.25 & 97.13 & 88.11 & \textbf{86.11} \\ 
\midrule \textbf{InfoPDF (Beta)} & & & & & & \\ 
\quad w/ GCN & 85.34 & 85.01 & 96.08 & 95.94 & 87.68 & 85.36 \\ 
\quad w/ GAT & 85.35 & 85.04 & 95.89 & 95.74 & 87.93 & 85.61 \\ 
\quad w/ SAGE & 85.34 & 85.03 & 96.17 & 96.01 & 87.41 & 85.15 \\ \bottomrule \end{tabular} }
\caption{Comparison results (\%) of InfoPDF using different distributions for reliability estimation. Best results are in \textbf{bold}.} \label{tab:beta} 
\end{table}

\begin{figure}[t]
    \centering
    \includegraphics[width=\linewidth]{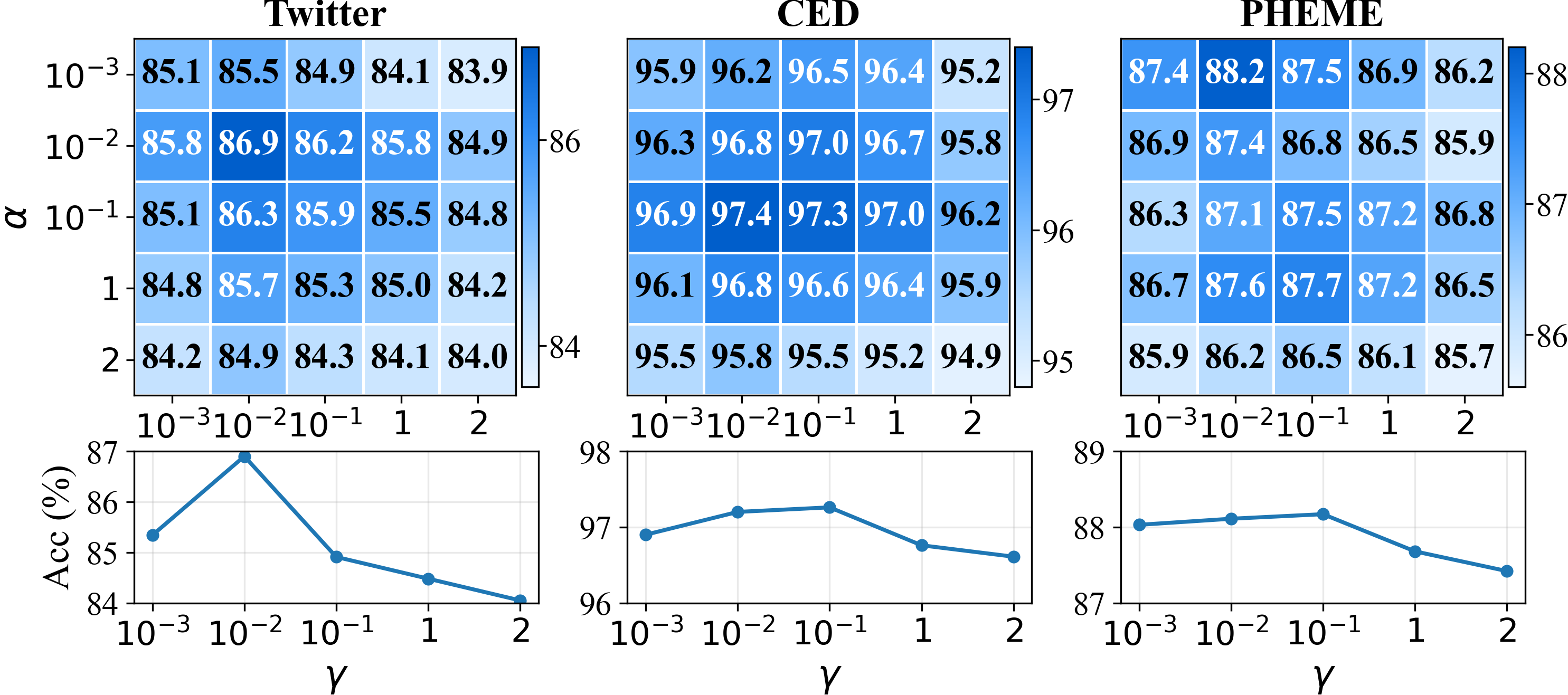}
    \caption{Performance analysis with varying $\alpha$, $\lambda$ and $\gamma$ parameters. We use accuracy score (\%) for evaluation. }
    \label{fig:para_analysis}
\end{figure}

We study reliability estimation with different probabilistic distributions. Besides the Gaussian distribution, we implement InfoPDF with a Beta variant by converting the logits produced by each attribute-specific probe into non-negative evidence and using the resulting uncertainty to compute reliability weights~\cite{han2022trusted,ma-etal-2024-event}. As shown in Table~\ref{tab:beta}, InfoPDF achieves competitive performance with both Gaussian and Beta distributions, demonstrating that our probabilistic reliability estimation remains effective under different distributional choices. These results further suggest that distributional uncertainty provides a useful signal for modeling the reliability of synthetic propagation views.

\paragraph{Hyperparameter Analysis.}\label{hyper}
We analyze the sensitivity of InfoPDF to $\alpha$, $\lambda$, and $\gamma$ in Figure~\ref{fig:para_analysis}.
A small $\lambda$ (e.g., 0.01) generally performs well, suggesting that mild alignment helps reduce the real--synthetic gap.
The optimal $\alpha$ varies across datasets, likely due to different attribute distributions and propagation patterns.
Moderate $\gamma$ values better balance denoising and information preservation, while excessive compression may remove task-relevant signals.
Overall, InfoPDF shows reasonable stability under different hyperparameter settings.

\section{Related Work}

\paragraph{Fake News Detection.} 
Existing methods primarily focus on content and propagation. 
Content-based methods extract semantic patterns via feature engineering~\cite{castillo2011information,ma2015detect}, deep neural networks~\cite{ruchansky2017csi,yu2017convolutional}, and PLMs~\cite{kaliyar2021fakebert}. Auxiliary tasks like sentiment analysis~\cite{luvembe2023dual} and comment utilization~\cite{shu2019defend} are also employed to complement textual features.
Propagation-based methods model dissemination structures using time series~\cite{Liu2018EarlyDO} or graphs~\cite{bian2020rumor,wei-etal-2021-towards,wei2022uncertainty}, sometimes incorporating multi-relational interactions~\cite{dou2021user}. 
However, incomplete propagation remains a challenge~\cite{wei-etal-2021-towards,ma2022towards}. While contrastive learning~\cite{ma2022towards,cui2024propagation} and uncertainty-aware modeling~\cite{wei-etal-2021-towards,wei2022uncertainty,lian2026decompose} improve robustness, existing propagation-based methods are still vulnerable to the incomplete propagation scenarios.

\paragraph{LLM-based Propagation Generation.} LLMs demonstrate significant potential in simulating human behavior and social knowledge~\cite{Liu2024FromST,wang2025collaboration}, having recently been established as a verifiable and scalable research method for capturing complex social dynamics~\cite{anthisposition}. Studies on fake news detection have moved beyond simple discussion simulation~\cite{nan2024let,wan2024dell} to modeling intricate dissemination behaviors such as liking, sharing, and content regulation within multi-agent social environments~\cite{Liu2025MOSAIC}. However, these methods often overlook the inherently unreliable risks in synthetic propagation. 
We mitigate the unreliability in LLM-generated propagation from the mutual information perspective.

\section{Conclusion}

This paper alleviates the unreliability in LLM-generated propagation from the mutual information perspective. We propose a novel InfoPDF to learn effective representations from both real and multiple LLM-generated synthetic propagation for fake news detection.
InfoPDF denoises unreliable synthetic propagation via a variational information bottleneck and maintains the consistency between real and synthetic propagation representations as well as the task sufficiency of fake news detection and user attribute prediction.
Experiments on three datasets demonstrate that InfoPDF achieves superior performance and robustness in data-sparse scenarios, while its LLM-free inference mode offers a practical solution for efficient deployment.

\section*{Acknowledgments}
This work was supported by the National Natural Science Foundation of China (No.U24A20335), the China Postdoctoral Science Foundation (No.2024M753481), and Youth
Innovation Promotion Association CAS. 

\bibliographystyle{named}
\bibliography{ijcai26}

\end{document}